# Toward Auditable Fraud Detection: Combining Graph Features, Model Explanations, and Agentic Case Investigation

Rahil Sharma
University of Amsterdam
Amsterdam, The Netherlands
rahil.sharma@student.uva.nl

## Abstract

Fraud detection systems must scale with rising transaction volume while remaining explainable and reviewable. This study examines a layered pipeline on the PaySim dataset combining a gradient-boosted classifier, graph-derived features, an autoencoder-based anomaly signal, TreeSHAP explanations, and a bounded LLM investigation agent applied to uncertain cases. After removing a simulator-specific balance shortcut that inflates baseline performance, neither the graph features nor the anomaly signal improves Average Precision on the full test set, though both rank fraud better among intermediate-score cases. Engineered structural features recover all transactions in a controlled multi-account fraud-ring experiment, while the tabular baseline misses roughly a quarter. The investigation agent underperforms direct classifier thresholding, 65.0% versus 71.7% accuracy on a balanced case sample; most disagreements replace correct classifier decisions with errors, each accompanied by a coherent rationale. Detection components contribute only under specific conditions, and a plausible agent rationale is not evidence of a better decision.

## Introduction

Financial institutions face a substantial volume of transaction alerts that require investigation. Industry reporting estimates that false positive alerts consume around 27% of anti financial crime team hours on average, while many institutions also report that their available resources are insufficient to address financial crime effectively (Nasdaq Verafin, 2026). At the same time, fraud techniques are becoming increasingly automated, with recent industry forecasts identifying the use of agentic AI by fraudsters as an emerging threat (Experian plc, 2026). These developments increase pressure on financial institutions to automate fraud analysis while preserving sufficient transparency, documentation, and human oversight.

Agentic systems are increasingly presented as tools for gathering evidence, summarising cases, and reducing the manual effort involved in alert review (Emburse, 2026; Nasdaq Verafin, 2026). Recent academic work has also begun to evaluate agent based approaches to fraud detection rather than treating them only as architectural proposals (Chen et al., 2026). However, it remains unclear whether an investigation agent, meaning an LLM-based agent that gathers evidence on an individual case and returns a verdict with a written rationale, improves decisions beyond the underlying classifier, and under what conditions its output should be accepted automatically or passed to a human reviewer. Throughout this paper, the terms investigation agent and LLM-based agent refer to this same component.

A related set of questions appears in the academic literature. Graph based fraud detection is a well established research area, with recent work addressing problems such as limited labels, heterophilous graph structure, and distribution shift (Cheng et al., 2025; Pan et al., 2025; Pan et al., 2026). Explainability has also become an important concern in fraud detection, including post hoc feature attribution for tabular models and graph based systems that incorporate explanation directly into the detection architecture (Li et al., 2024; Lundberg et al., 2020). Recent work has further begun to connect large language models with fraud detection and graph based reasoning, including LLM enhanced graph fraud detection (Huang et al., 2025), relational soft prompting (Zuo et al., 2026), reasoning distillation for graph detectors (Li et al., 2026), and multi agent fraud detection frameworks (Chen et al., 2026). Broader work on graph augmented

LLM agents provides additional context for how graphs can support planning, memory, and tool use in agent systems (Liu et al., 2026).

These studies show that graph information, explanation, and LLM based reasoning are already beginning to converge. What remains less clear is how their contributions should be separated within a shared fraud analysis pipeline. In particular, there is limited evidence on whether a bounded investigation agent, one restricted to a fixed set of predefined evidence tools and a capped number of interaction rounds, adds decision value beyond the classifier on which it relies, how graph and anomaly signals behave specifically on difficult cases, and how disagreement between the classifier and the agent can be used to decide when human review is required.

This work examines how the different components of an automated fraud analysis system can support selective human review while preserving transparency and auditability. The study focuses on two questions. The first asks whether graph derived structural features and an autoencoder based anomaly signal provide additional detection value, particularly among cases that receive intermediate scores from a supervised baseline. The second asks whether a bounded agentic investigation layer can use model explanations, graph context, and retrieved reference cases to produce structured evidence for these difficult cases, and whether disagreement between the agent and the classifier can help identify cases that should be escalated to a human reviewer.

Figure 1 summarises the overall approach, beginning with the operational need for scalable and reviewable fraud detection and ending with the policy used to route cases between automatic resolution and human review. In this study, auditability refers to the ability to reconstruct which model outputs, retrieved evidence, and decision rules contributed to a case outcome. It does not refer to formal regulatory compliance or to independently validated explanation quality.

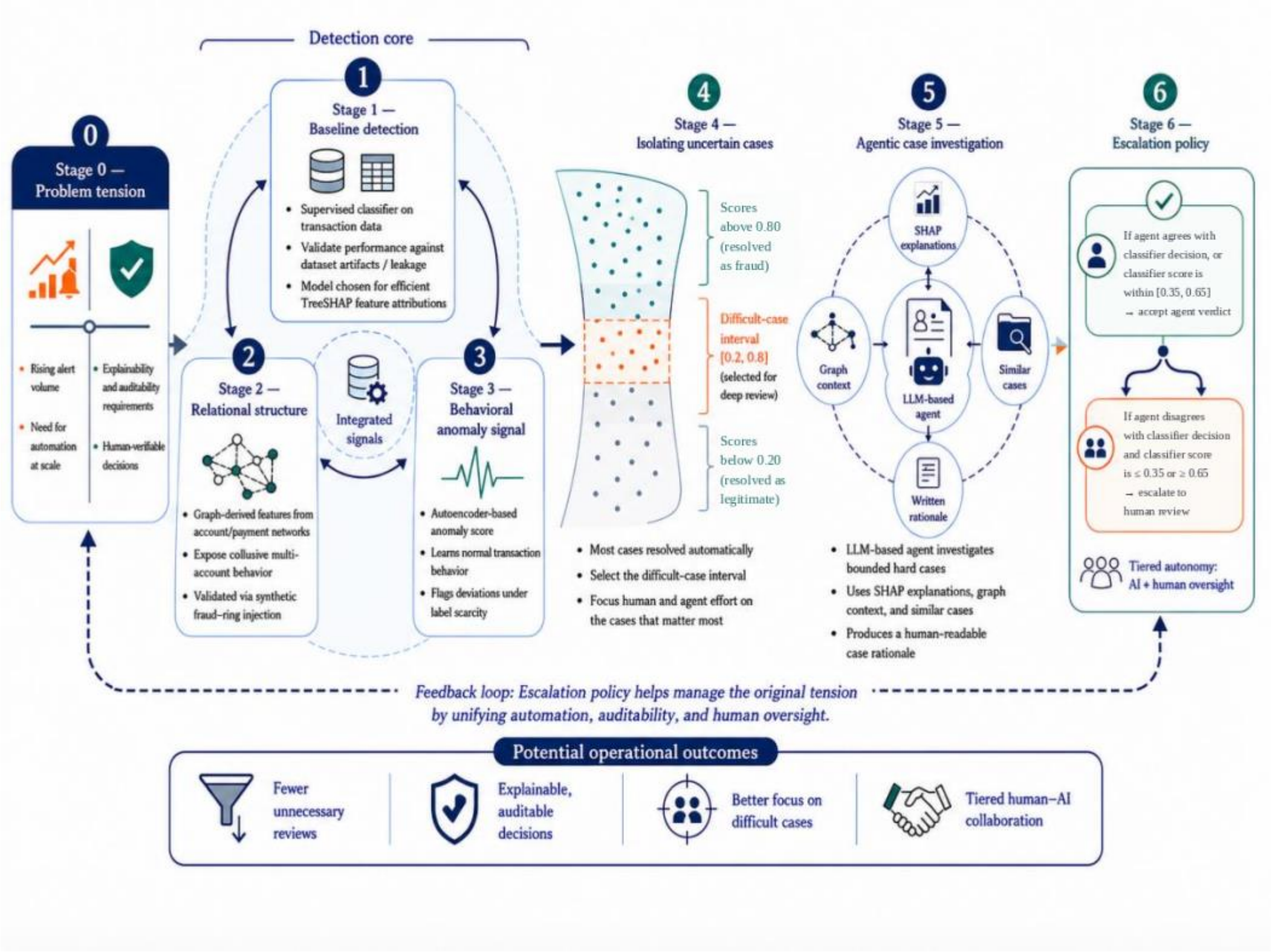

**Figure 1. Overview of the proposed framework, from the operational tension motivating this work (Stage 0) through baseline detection, relational and behavioral feature augmentation, difficult-case isolation, agentic investigation, and the escalation policy that routes cases between automatic resolution and human review. All scores in Stages 4 and 6 refer to the output of the gradient-boosted classifier.**

## Related Work

Fraud detection is commonly approached as a supervised classification problem in which a model learns from labelled historical transactions and distinguishes fraudulent from legitimate activity using attributes such as amount, timing, transaction type, and account behaviour. This representation is effective when fraud can be identified from the characteristics of an individual transaction, but it is less suited to patterns that depend on relationships between multiple entities or events. Graph based fraud detection addresses this limitation by representing accounts, devices, merchants, or transactions as nodes and their interactions as edges (Cheng et al., 2025; Pourhabibi et al., 2020). This relational view is particularly relevant to coordinated behaviour such as fraud rings, funnel accounts, and chains of transfers. An individual transaction may appear ordinary when considered alone, while the wider pattern of accounts and transfers around it may provide important evidence of coordination. Such information is not present in the transaction's original attributes and must be obtained from relationships across the wider network.

Relational information can be incorporated into fraud detection in several ways. The simplest approach is to derive structural statistics such as node degree, the number of distinct counterparties, or measures of centrality and then provide these values as features to a conventional classifier. A second approach uses graph embedding methods such as node2vec, which generates biased random walks through a network and applies a Skip-gram objective so that nodes appearing in similar walk contexts receive similar numerical representations (Grover & Leskovec, 2016). A third approach uses graph neural networks, which learn node representations by repeatedly combining a node's current representation with information from its neighbours (Kipf & Welling, 2017). These approaches differ in complexity and expressive power. Engineered graph features and standalone embeddings can be integrated into standard tabular models, while end-to-end GNNs learn the relational representation and prediction function jointly.

Graph based approaches to fraud and financial crime detection are now a well established area of research, with recent work continuing to address challenges including severe label scarcity and changes in transaction patterns that can reduce the effectiveness of models trained on historical graph data (Pan et al., 2025; Pan et al., 2026). Similar ideas are also used in industry systems; PayPal has described a real time graph platform that computes structural information from relationships between accounts and transactions and makes this information available to downstream risk systems (PayPal Engineering, 2021). This work follows the same general strategy by supplying engineered graph features and node2vec embeddings to a downstream classifier, generating learned embeddings only for the subgraph containing accounts that appear in more than one transaction, reflecting the structure of the PaySim data where most accounts occur only once.

A second concern in modern fraud detection is the ability to explain why a model produced a particular prediction. In this work, explanation refers to a local account of how individual input features contributed to a specific model output, rather than to a general description of the model as a whole. SHAP is a widely used method for producing this kind of feature attribution in tabular machine learning, based on Shapley values from cooperative game theory (Lundberg & Lee, 2017). For tree based models, TreeSHAP exploits the structure of the trees to compute exact SHAP values efficiently under a specified explanation formulation (Lundberg et al., 2020). For deep neural networks and graph neural networks, practical explanation methods more commonly rely on approximation or on methods designed specifically for the model and data structure (Yuan et al., 2023).

Gradient boosted decision tree ensembles construct a predictive model as an additive sequence of decision trees, allowing the model to capture complex relationships while retaining the structure of a tree based predictor (Friedman, 2001). XGBoost is an efficient and regularised implementation of this approach and is the implementation used in this work (Chen & Guestrin, 2016). Tree based models are particularly useful for the present study because they can be combined with TreeSHAP to produce local feature attributions efficiently (Lundberg et al., 2020), allowing graph derived and anomaly based signals to be added as ordinary input features while preserving a consistent explanation method across all model variants. An end-to-end graph neural network would instead learn directly from the graph and would require a different family of explanation methods, making it more difficult to separate the value of relational information from the effect of changing the underlying classifier. The use of XGBoost is therefore a design choice motivated by the need to balance predictive modelling, component level comparison, and reviewable feature attribution.

A more recent development is the use of large language models as components within fraud detection and investigation systems. In this work, an agentic system refers to a language model that can select from a bounded set of tools, retrieve case specific evidence, and use the retrieved information before producing a structured decision. In the present study the set contains four evidence tools, returning transaction details, the classifier score with its largest SHAP attributions, graph context for the accounts involved, and distances to labelled reference cases, together with a verdict tool that ends the investigation (see the Method section). Applied to fraud investigation, this creates a different role from that of the underlying classifier: rather than producing a score from a fixed set of features, the agent can inspect transaction details, model explanations, graph context, and similar historical cases before returning a verdict and written rationale. Recent research has begun to connect LLMs with graph based fraud detection in several ways, including using LLM derived semantic information to improve graph representations (Huang et al., 2025), translating relational structure into soft prompts for joint LLM and GNN modelling (Zuo et al., 2026), distilling graph aware reasoning into a fraud detector (Li et al., 2026), and multi-agent frameworks for constructing and refining broader fraud detection workflows (Chen et al., 2026). Closely related, FLAG selects graph neighbours by the semantic similarity of their text and uses an LLM to extract discriminative textual features for each node, so that a graph neural network on text-rich graphs is less affected by fraudsters camouflaging themselves among legitimate neighbours (Yang et al., 2025), while Luo et al. (2025) combine semantic features extracted with GPT-4o with a graph convolutional network over a consumer and merchant transaction graph for e-commerce payment fraud. These methods mainly use LLMs to improve the detection model itself, and they are evaluated on whether detection performance improves. The role examined in this work is narrower: the LLM does not replace or train the fraud classifier, but operates after the classifier and investigates only a bounded set of cases using evidence made available through predefined tools. This design reflects an operational model in which clear cases may be handled automatically while uncertain or conflicting cases are passed to a human reviewer (Emburse, 2026; Nasdaq Verafin, 2026). The contribution of the present work is therefore complementary rather than competing: instead of asking whether an LLM can raise detection accuracy, it asks whether an LLM placed after a fixed classifier adds decision value, and it evaluates the reviewability of the agent's rationale separately from the correctness of its verdict.

Taken together, these studies show that graph-based detection, model explanation, and LLM-based reasoning have historically represented distinct parts of the fraud analysis problem but are now beginning to converge. Production systems also illustrate that these components may be combined in different ways depending on operational priorities. PayPal has described the use of graph-derived features within downstream risk models and a real-time graph platform designed to support fraud detection at scale (PayPal Engineering, 2021), though the reviewed public architecture does not describe a downstream LLM investigation layer of the kind examined in this work. Third-party reporting on PayPal's recent fraud prevention approach also emphasises transaction time risk assessment and proactive intervention (Chief AI Officer, 2026). Stripe Radar provides a further example of how production fraud systems are shaped by scale, latency, feature availability, and the need to update models rapidly (Stripe, n.d.). Industry commentary suggests that many current agentic applications in financial services remain closer to decision support, case analysis, and workflow assistance than to fully autonomous investigation (Zelezkins, 2026). Within this developing landscape, the present work focuses on a narrower question: it evaluates the contribution of graph-derived and anomaly-based signals relative to a leakage-checked tabular baseline, both across the full test set and within a fixed subset of difficult cases, and evaluates a bounded LLM agent that operates after classification, comparing its verdicts directly with the classifier on which it relies.

# Method

Figure 2 presents the full pipeline used in this study. Transaction data is processed through three feature sources before classification. Cases receiving intermediate classifier scores are then passed to the bounded agentic investigation layer and the subsequent escalation policy.

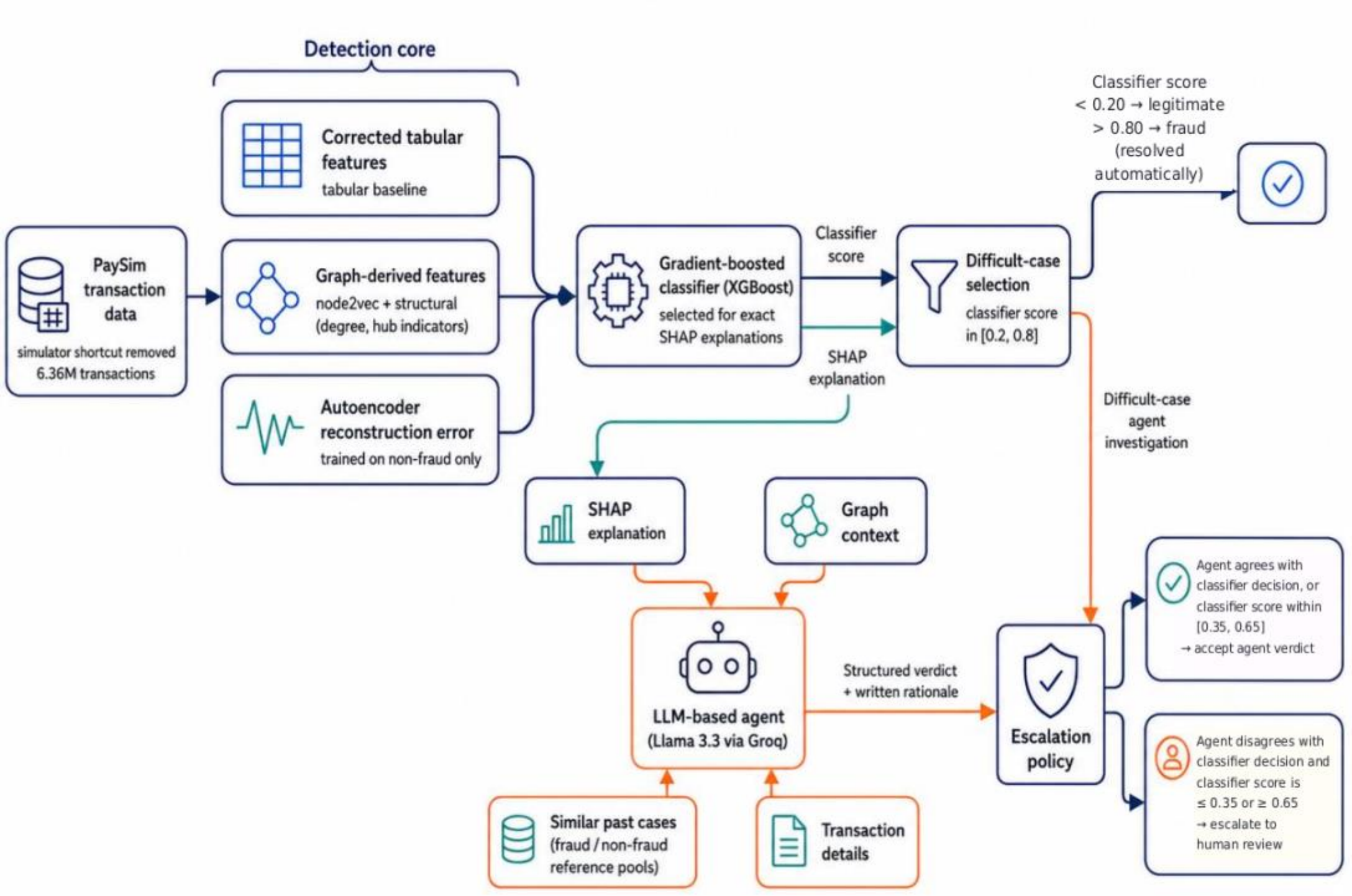

**Figure 2. System architecture of the fraud detection and investigation pipeline. Transaction data is represented through corrected tabular features, graph-derived features, and an autoencoder-based anomaly signal. These inputs are supplied to a gradient-boosted classifier, which produces a classifier score and a SHAP feature attribution. Cases receiving scores within the selected difficult-case interval are passed to a bounded LLM investigation agent with access to transaction details, graph context, model explanations, and retrieved reference cases. The agent's verdict is then evaluated by an escalation policy that determines whether the case is resolved automatically or passed to human review.**

## *Data*

This work uses PaySim, a synthetic mobile money transaction dataset generated from aggregated patterns derived from a real mobile money service (Lopez-Rojas et al., 2016). The public dataset contains 6,362,620 transactions across five transaction types: CASH_IN, CASH_OUT, DEBIT, PAYMENT, and TRANSFER. PaySim was selected because it contains persistent sender and recipient account identifiers, allowing an account-level transaction graph to be constructed directly from the recorded transfers. The dataset is highly imbalanced, and fraudulent transactions occur only within the TRANSFER and CASH_OUT categories. All five transaction types are retained rather than filtering the data to these two categories, which prevents the evaluation from assuming in advance which transaction types can contain fraud.

Table 1 lists the fields in PaySim, their meaning, and how each is used in this study. Each record describes one transaction by its time step, where one step corresponds to one hour of simulated time over 743 steps, its type and amount, the sender and recipient account identifiers, and the balances of both accounts before and after the transaction. Of the 6,362,620 transactions, 8,213 (0.129%) are labelled fraudulent, comprising 4,097 TRANSFER and 4,116 CASH_OUT transactions, which corresponds to a fraud rate of 0.77% among TRANSFER transactions and 0.18% among CASH_OUT transactions; the other three types contain no fraud. The dataset also includes a rule-based flag, isFlaggedFraud, which is set for only 16 transactions and reflects a simulator business rule rather than an observable transaction attribute, so it is excluded. The account identifiers are not used as model inputs but define the transaction graph described below. After one-hot encoding of the transaction type, the corrected tabular feature set contains ten features.

**Table 1. PaySim fields and their use in this study. The ten model inputs form the corrected tabular feature set.**

| Field | Description | Use in this study |
|---|---|---|
| step | Hour of simulated time (1 to 743) | Model input |
| type | CASH_IN, CASH_OUT, DEBIT, PAYMENT, TRANSFER | Model input (one-hot, five columns) |
| amount | Transaction amount | Model input |
| nameOrig, nameDest | Sender and recipient account identifiers | Graph construction only |
| oldbalanceOrg, newbalanceOrig | Sender balance before and after | Excluded (simulator shortcut) |
| oldbalanceDest, newbalanceDest | Recipient balance before and after | Model input |
| errorBalanceDest (derived) | newbalanceDest minus (oldbalanceDest + amount) | Model input |
| isFlaggedFraud | Simulator rule-based flag (16 transactions) | Excluded |
| isFraud | Ground-truth fraud label | Prediction target |

During exploratory modelling, a feature measuring the difference between the recorded origin account balance change and the transaction amount produced unexpectedly strong predictive performance, and with the origin balance fields included the baseline reached an Average Precision of 0.998. Further analysis showed that the origin side balance arithmetic reconciled almost exactly for fraudulent transactions far more often than for legitimate transactions: the discrepancy was below one currency unit in absolute value for 99.45% of fraudulent transactions but only 21.3% of legitimate ones. This relationship is unlikely to represent a general property of fraudulent behaviour and is instead consistent with a shortcut introduced by PaySim's simulation process. Removing only the derived discrepancy feature was not sufficient because a tree based model could reconstruct the same relationship from the origin balance fields and amount. The two origin balance fields were therefore excluded from the reported baseline, while amount was retained as an ordinary transaction attribute. The destination balance fields, and the analogous destination discrepancy feature derived from them, did not exhibit the same strong and nearly deterministic relationship with the fraud label and were therefore retained. This correction was applied before the model comparisons reported in the remainder of the study.

## *Graph Construction and Feature Generation*

The transaction graph is constructed by treating each account identifier as a node and each transaction as a directed edge from the sending account to the receiving account. Examination of the resulting network revealed a strong asymmetry between sender and recipient behaviour: most sending accounts occur in only one transaction, while a smaller group of recipient accounts appear repeatedly and receive transfers from multiple distinct senders. Two forms of graph based representation are derived from this structure. The first consists of engineered structural features: the total number of times the origin and destination accounts appear in the dataset, indicators showing whether each account appears more than once, and the number of distinct origin accounts connected to the destination. The second consists of node2vec embeddings learned from the subgraph induced by accounts that appear in more than one transaction (Grover & Leskovec, 2016), with an embedding dimension of 32 for each account, giving separate origin and destination representations. Accounts outside this recurring subgraph receive a zero vector because their limited neighbourhood structure provides little information for representation learning.

All structural features and node2vec embeddings are generated from the complete transaction graph before the random train and test split is applied. The resulting evaluation is therefore transductive: fraud labels from the test set are not used during feature generation, but the graph representations can include relationships involving transactions that are subsequently assigned to the test partition. The reported graph results should consequently be interpreted as performance when the complete transaction network is available, rather than as a fully inductive simulation in which future transactions are unseen.

### *Anomaly Signal*

A secondary anomaly signal is generated using a small autoencoder trained on legitimate transactions from the training set, following prior use of autoencoders and other unsupervised models for transaction fraud detection (Misra et al., 2020; Niu et al., 2019), and using the same corrected tabular feature set as the baseline classifier. The autoencoder has a symmetric encoder and decoder with layer dimensions d to 8 to 4 to 8 to d, where d = 10 is the number of input features listed in Table 1, using ReLU activations and trained for 20 epochs using mean squared reconstruction loss. For each transaction, the anomaly signal is defined as the mean squared error between the standardised input features and their reconstruction, and this reconstruction error is added as a single input feature to the gradient-boosted classifier described in the next subsection, rather than being used as a standalone fraud decision. The autoencoder does not use fraudulent examples to learn a direct fraud decision boundary; however, known training labels are used to restrict autoencoder training to legitimate transactions, so this component should be understood as a one-class anomaly signal rather than a fully label-free method.

### *Detection Model*

The primary detection model is an XGBoost gradient boosted decision tree ensemble (Chen & Guestrin, 2016). The same model configuration is used across the baseline and feature augmented variants so that differences in performance can be attributed to the feature sets rather than to changes in the classifier. Each model contains 300 trees with a maximum depth of 6 and a learning rate of 0.1, with a fixed random seed of 42 used throughout. To account for the strong class imbalance, the positive class weight is set to the ratio between the number of legitimate and fraudulent transactions in the training set. XGBoost was selected because it provides a strong model for heterogeneous tabular features while remaining compatible with TreeSHAP feature attribution (Lundberg et al., 2020), allowing the same explanation method to be used for the baseline, graph augmented, autoencoder augmented, and combined models.

No explicit interaction terms are constructed. The graph-derived features and the reconstruction error enter the classifier as additional columns alongside the corrected tabular features, and any interaction among them is learned through the tree structure, where a split on one feature conditions the subsequent splits on others. Because all variants share the same model configuration and TreeSHAP attribution, the contribution of each added feature can be compared directly. In the graph-augmented model, which has 79 input features, mean absolute SHAP values on a random sample of 2,000 test transactions ranked the number of distinct senders to the destination account ninth and the destination account's degree eleventh, while origin degree and both repeat indicators received zero attribution, consistent with most sending accounts appearing only once. The 64 node2vec dimensions contributed little individually: no dimension ranked above twelfth, and the largest mean absolute SHAP value among them was 0.013, compared with 0.40 for the distinct-sender count.

### *Agentic Investigation Layer*

The agentic investigation layer is evaluated only on transactions for which the autoencoder-augmented classifier, the variant with the highest Average Precision within the difficult subset, produces a score between 0.2 and 0.8, used as a practical definition of cases requiring further investigation. Because a random sample from this subset would contain very few fraudulent transactions, the evaluation set is constructed as a balanced case-control sample containing 30 fraudulent and 30 legitimate transactions. The true labels are used only to construct and evaluate this sample and are not shown to the agent.

The investigation agent uses Llama 3.3 70B through the Groq inference service with a temperature of 0.1. For each case, the agent receives only a case identifier and can retrieve evidence through four predefined tools, providing transaction details, the classifier score and five largest SHAP feature attributions, graph context for the origin and destination accounts, and feature-space distances to labelled fraud and legitimate reference examples from the training set. The reference retrieval tool maintains separate fraud and legitimate pools and returns distances to the two nearest examples from each pool, with features standardised before Euclidean distances are computed. This separation prevents the much larger legitimate class from dominating a single nearest-neighbour search.

The agent is instructed to inspect at least two evidence sources before reaching a conclusion, may perform up to eight interaction rounds, where one round consists of a single call to the language model together with the execution of any tool requests in its response, and must finish by calling a structured verdict tool with one of three outputs: fraud, not fraud, or escalate. The verdict also includes a written rationale that refers to the evidence retrieved during the investigation. All tool calls, returned evidence, and the final rationale are stored as a reviewable case transcript, used in this study to inspect the agent's reasoning and identify recurring failure patterns, though not treated as a formal validation of audit trail quality.

# Experimental Setup

## *Model Comparisons*

Four model variants are evaluated using the same training and test transactions: a hard baseline using only the corrected tabular feature set; a graph-augmented variant adding the engineered structural features and node2vec derived features; an autoencoder-augmented variant adding the reconstruction error; and a combined variant including both additions. The same XGBoost configuration is used for all four variants. A single random 80/20 train and test split, stratified by the fraud label with a fixed random seed of 42, is used across all variants. The split is not temporal, and the graph statistics and node2vec representations are generated from the complete transaction graph before the split, so the graph-based evaluation follows the transductive setting described above.

A separate controlled experiment examines how the tabular and graph-augmented models respond when a coordinated multi-account pattern is introduced into PaySim. Forty synthetic rings are generated, each containing five accounts and six fraudulent transfer transactions, with amounts and starting steps sampled from the PaySim distribution and account balances updated consistently to avoid reproducing the origin balance shortcut removed from the baseline. Individual transactions, rather than entire rings, are randomly assigned to the two partitions, so the experiment should be interpreted as a controlled test of whether the available structural features help identify further transactions belonging to an introduced recurring pattern, not as a test of generalisation to entirely unseen fraud rings.

## *Evaluation Metrics*

Because fraudulent transactions form a very small fraction of PaySim, classification accuracy is not used as the primary metric. The primary ranking metric is Average Precision, computed using the predicted fraud scores across all decision thresholds, which is more informative than accuracy or a ROC curve when the main concern is retrieving a rare positive class (Saito & Rehmsmeier, 2015). Performance is evaluated both on the complete test set and on a fixed subset of cases receiving intermediate scores from the hard baseline (between 0.2 and 0.8 inclusive), used as a proxy for cases that are more difficult for the baseline to separate. Because this subset is selected using the hard baseline's own scores, the comparison is intentionally conditional on baseline uncertainty and should not be interpreted as an unbiased subgroup comparison across all model variants.

For the agentic investigation layer, the final verdict is compared with the ground truth label, with escalate scored as incorrect for the raw verdict accuracy since it does not produce a binary classification. Agent performance is compared with a classifier decision baseline that labels a transaction as fraud when the underlying model score is at least 0.5 and as legitimate otherwise, testing whether the agent improves the binary decision beyond directly thresholding the same classifier score available to it. Cases where the agent's binary verdict differs from the thresholded classifier decision are examined separately, recording how often these changes correct a classifier error and how often they introduce a new error.

# Results

### *Detection Performance*

Figure 3 and Table 2 report Average Precision on the complete test set and on the baseline-defined difficult subset. The difficult subset contains 103,165 transactions, representing 8.11% of the test set, including 184 fraudulent transactions (approximately 0.18% of this subset). Although the augmented variants obtain higher observed ranking performance than the hard baseline within this subset, the absolute Average Precision remains low, indicating that these transactions remain difficult to separate reliably.

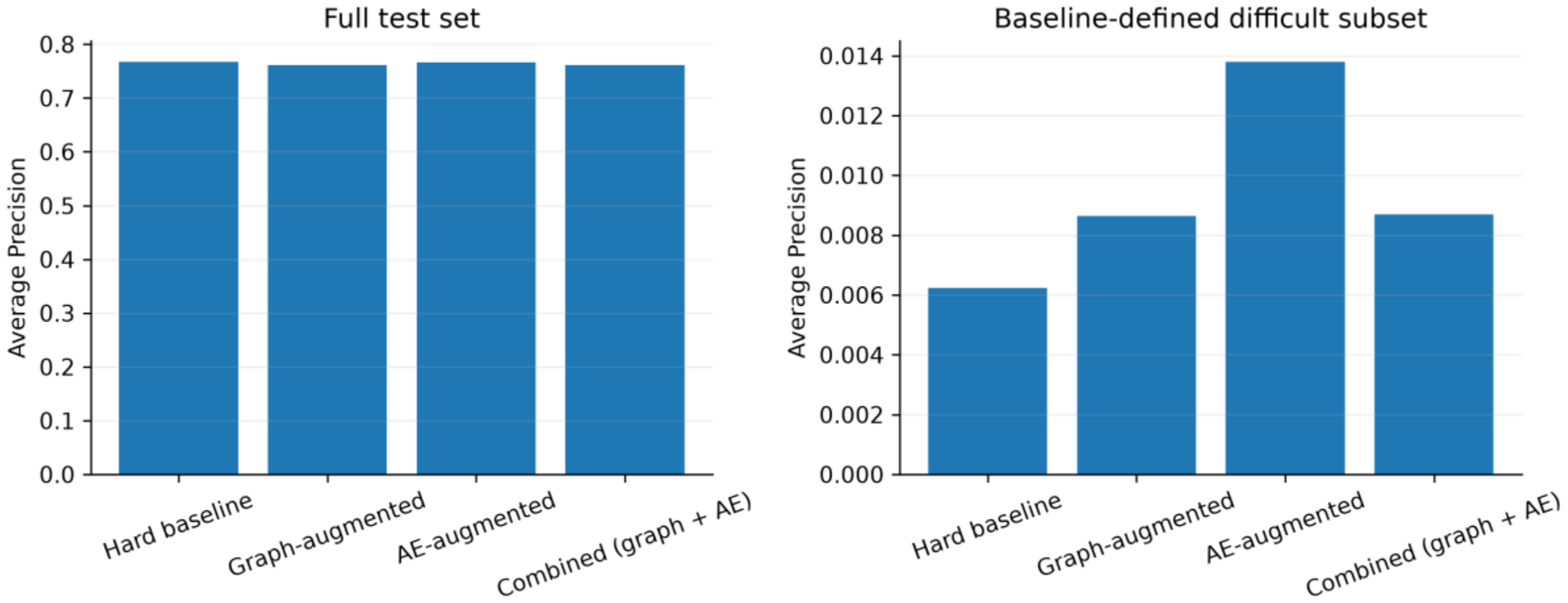

**Figure 3. Average Precision across model variants on the complete test set and the baseline-defined difficult subset. The hard baseline obtains the highest observed value on the complete test set, while the autoencoder-augmented variant obtains the highest value within the difficult subset. The two panels use different vertical scales.**

**Table 2. Average Precision across model variants on the complete test set and the baseline-defined difficult subset.**

| Variant | Full AP | Difficult AP |
|---|---|---|
| Hard baseline | 0.7677 | 0.0062 |
| Graph augmented | 0.7611 | 0.0086 |
| Autoencoder augmented | 0.7662 | 0.0138 |
| Combined | 0.7617 | 0.0087 |

Within the baseline-defined difficult subset, the autoencoder-augmented model obtains the highest observed Average Precision, followed by the combined and graph-augmented variants. The combined model does not preserve the improvement observed for the autoencoder-only model. One possible explanation is that the additional graph representation introduces many low-contribution dimensions, including 64 node2vec features, while the subset contains only 184 fraudulent transactions; however, the present experiment does not isolate the cause of this difference. The result should therefore be interpreted as evidence that combining additional feature sources does not necessarily improve performance, rather than as a demonstrated effect of dimensionality.

### *Controlled Multi-Account Pattern Experiment*

A separate experiment examines whether relational features contribute when a recurring multi-account fraud pattern is introduced into PaySim. Thirty-nine of the injected transactions occurred in the test partition. The hard baseline detected 30 of the 39 injected test transactions, giving a recall of 76.9%. The graph-augmented model detected all 39 transactions. Recall on the original PaySim fraud cases was 0.8923 for the hard baseline and 0.8838 for the graph-augmented model. The graph features therefore improved recall on the introduced multi-account pattern while producing a small reduction on the original PaySim fraud cases. An analysis using the true generated hop position found no consistent decline in the hard baseline score as transactions moved through the rings; missed transactions occurred at several different positions, and the mean baseline score varied irregularly across hops. The broader result is that structural account features helped identify a recurring multi-account pattern that the transaction-level baseline did not detect consistently.

The experiment remains transductive because individual transactions from the same ring may appear in both training and test partitions, and graph statistics are computed from the complete merged graph. In addition, the injected accounts are absent from the precomputed node2vec vocabulary and receive zero embeddings, so the observed improvement should be attributed mainly to the engineered structural features.

### *Agent Decisions and Escalation*

The bounded investigation agent was evaluated on the 60 cases described above. The agent produced a correct binary verdict in 39 cases, giving an accuracy of 65.0%. Directly thresholding the same classifier score at 0.5 produced 43 correct decisions, giving an accuracy of 71.7%. The agent therefore did not improve the underlying classifier decision on this evaluation set. The agent gave a binary verdict that differed from the classifier decision in eight cases; only two of these changes corrected a classifier error, while the remaining six replaced a correct classifier decision with an incorrect agent verdict. This result is operationally important because the agent often produced a coherent written rationale even when changing the decision in the wrong direction; a plausible explanation alone was therefore not a reliable indication that the agent had improved the fraud judgement.

Manual inspection of the transcripts identified a recurring failure pattern: the agent sometimes treated the relative distance to labelled fraud and legitimate reference examples as decisive evidence, even when the classifier score pointed in the opposite direction, despite instructions to treat the classifier output as the strongest single signal.

**Table 3. Classifier and agent decisions on the balanced 60-case evaluation sample. These values are comparative results on a case-control sample and are not deployment performance estimates.**

| Measure | Classifier | Agent |
|---|---|---|
| Correct decisions | 43/60 | 39/60 |
| Accuracy | 71.7% | 65.0% |

Based on the observed failure pattern, an exploratory escalation rule was applied after the agent evaluation: a case was passed to human review when the agent disagreed with the thresholded classifier decision and the classifier score lay outside the interval [0.35, 0.65], that is, at most 0.35 or at least 0.65. Cases in which the agent itself returned escalate were likewise passed to human review. All other cases, namely those where the agent agreed with the classifier or where the classifier score lay within [0.35, 0.65], were resolved automatically using the agent's verdict. Under this rule, two of the 60 cases were escalated, corresponding to an escalation rate of 3.3% and automatic-decision coverage of 96.7%. Both escalated cases were agent errors, and no correct agent decisions were escalated in this sample. Among the remaining 58 cases, 39 decisions were correct, giving a selective accuracy of 67.2%. The rule was developed after inspecting the same cases on which it was evaluated, so its performance should be treated as exploratory rather than as evidence of a validated deployment policy.

### *Illustrative Case*

Transaction case_3929213 illustrates how a difficult case moves through the framework. It is a CASH_OUT of 411,566.70, in the 94th percentile of amounts for its type, sent to a destination account with a zero opening balance that had received transfers from nine distinct senders. The classifier assigned it a fraud probability of 0.677, placing it inside the difficult-case interval, and its two largest SHAP attributions, the destination's opening balance and the autoencoder reconstruction error, both pointed toward fraud. As in all 60 investigations, the agent retrieved the four evidence sources in a fixed order, namely transaction details, the classifier score with its SHAP attributions, graph context, and reference-case distances, before calling the verdict tool. Its rationale described the classifier score as "the strongest piece of evidence" and noted that the destination's nine senders "could suggest a suspicious pattern", but it concluded not fraud because the nearest legitimate reference case (distance 0.014) was closer than the nearest fraudulent one (0.072). The transaction was in fact fraudulent. Because the agent disagreed with a classifier score above 0.65, the escalation rule routed the case to human review, and the stored transcript allows a reviewer to see exactly which evidence the agent treated as decisive. This pattern was consistent across the sample: in all eight disagreements, the classifier score was above 0.5, the agent returned not fraud, and its rationale cited a smaller distance to the legitimate reference pool. The agent did not use the escalate option in any of the 60 investigations, which is why an explicit escalation rule, rather than the agent's own judgement, was needed to route such cases to human review.

# Discussion

The results suggest that the components of a layered fraud analysis system should not be expected to improve performance uniformly across all transactions. On the complete test set, neither the graph-based additions nor the autoencoder signal improved Average Precision over the corrected tabular baseline. Their value appeared instead in more specific settings: the autoencoder-based signal produced the highest Average Precision within the fixed subset of cases receiving intermediate baseline scores, while the controlled ring experiment showed that engineered graph features became useful when the data contained recurring multi-account structure. These findings support a conditional interpretation of additional feature sources. An anomaly signal may be most useful when a transaction differs from patterns learned from legitimate activity, while graph-derived features are most relevant when fraud depends on relationships across accounts rather than on the attributes of a single transaction. Adding such signals indiscriminately does not guarantee improvement and may introduce complexity without corresponding detection benefit.

This distinction is important for practical fraud system design. Aggregate performance remains necessary because most transactions pass through the primary detection model, but aggregate metrics can conceal differences within the smaller set of cases that require additional investigation. A production system would need to define investigation queues using operational thresholds, expected loss, review capacity, and the consequences of false positives and false negatives.

The agent evaluation provides a separate lesson. The bounded investigation agent had access to more contextual evidence than the thresholded classifier, including feature attributions, graph context, and retrieved reference distances, but it produced fewer correct verdicts. In most disagreements, the agent replaced a correct classifier decision with an incorrect one. This shows that access to additional evidence and the ability to generate a coherent rationale do not by themselves establish better decision quality. The written transcripts remained useful for identifying how the agent reached its conclusions, particularly its tendency to assign excessive importance to reference-case distances. However, reviewability should not be confused with correctness: an auditable error remains an error, even when its reasoning process can be inspected.

The more defensible role for the agent in the present architecture is therefore as a bounded investigation and documentation layer rather than as an autonomous replacement for the classifier. Such an agent could gather case evidence, organise model outputs, and prepare a structured record for review, while decision authority remains constrained by explicit system policies. The agent may still contribute operational value even when its verdict is not accepted directly, provided that the evidence it gathers is accurate, traceable, and useful to a human reviewer. Establishing that value would require evaluation with fraud analysts or other intended users, which is outside the scope of the present study.

The exploratory escalation rule illustrates both the usefulness and the limitations of explicit decision boundaries. In the evaluated sample, classifier-agent disagreement combined with a sufficiently strong classifier score identified two agent errors without escalating a correct agent decision, suggesting that disagreement can serve as one observable signal for human review. However, the rule was derived and assessed on the same small sample, and it addressed only cases where the agent overrode a comparatively decisive classifier output. Escalation should therefore be understood as a selective risk-control mechanism rather than as a complete solution to agent unreliability. More broadly, operational governance in fraud detection depends on how model outputs, evidence collection, decision rules, and human review are connected. These mechanisms do not constitute a complete governance framework, nor do they guarantee correct decisions; their value lies in making system behaviour easier to examine and in limiting the circumstances under which an automated verdict is accepted.

### *Implications for Research*

These findings speak to a broader tension in the human-AI collaboration literature between the appeal of algorithmic transparency and its actual evidentiary value. Prior work on automation reliance has shown that people's trust in a system does not always track its accuracy, and that both excessive aversion to algorithmic judgement and excessive appreciation of it can lead to poor reliance decisions (Dietvorst et al., 2015; Logg et al., 2019; Parasuraman & Riley, 1997). The present study extends this line of reasoning to a setting where the object of potential over-reliance is not the classifier itself but an agent's narrative justification for departing from it. A coherent rationale functioned, in this evaluation, as a source of misplaced confidence rather than as evidence of improved judgement. This suggests a research opportunity in extending trust-in-automation frameworks to explicitly separate reliance on a model's output from reliance on a model's explanation of that output, since the present results indicate these can diverge in their reliability even when produced by the same underlying system.

### *Implications for Practice*

For practitioners designing agentic layers for fraud review, the central implication is that a written rationale should not be treated as a proxy for decision quality during system validation. Organisations piloting LLM-based investigation agents should benchmark agent verdicts directly against the simplest available baseline, such as thresholding the underlying classifier, before treating the agent as additive. Escalation policies should be built on observable disagreement and confidence signals rather than on the perceived persuasiveness of the agent's explanation, since the two were shown here to be independent. Finally, the auditability provided by stored transcripts and feature attributions supports governance and post hoc review, but it does not by itself substitute for an accuracy evaluation; organisations should treat reviewability and correctness as separate design requirements to be validated independently.

## Limitations

The findings should be interpreted in light of several limitations. First, PaySim is a synthetic dataset whose fraud cases do not reproduce the full range of behaviours found in operational financial systems; its native fraud is closer to single-account takeover than to coordinated multi-account activity, which limits the extent to which the graph results can be generalised to real collusive fraud. Second, the model evaluation is based on a single random stratified train and test split rather than a temporal evaluation, and the graph evaluation is transductive because structural features and node2vec representations are generated from the complete transaction graph before the split. The reported graph results should therefore not be interpreted as fully inductive performance on previously unseen transactions or accounts.

The controlled ring experiment has related constraints: individual transactions rather than complete rings are assigned to the train and test partitions, so the experiment evaluates whether the model can identify further transactions belonging to an introduced recurring pattern, not whether it can generalise to an entirely unseen ring. The baseline-defined difficult subset contains 184 fraudulent transactions, and the higher Average Precision observed for the graph and autoencoder variants within this subset is based on a relatively small positive-class sample and one train and test split, with no confidence intervals, repeated splits, or statistical significance tests reported.

The agentic evaluation is smaller still, containing 60 cases selected to include equal numbers of fraudulent and legitimate transactions rather than reflecting the natural class distribution; the reported agent and classifier accuracies therefore describe comparative behaviour on this case-control sample and are not deployment performance estimates. The study also evaluates one run of one language model with a single prompt, tool configuration, and retrieval design, so the observed tendency to over-weight reference-case distances should be treated as a failure pattern of the evaluated configuration rather than a general property of all LLM-based fraud investigation agents. The escalation rule is exploratory and was developed after inspecting the same 60 cases on which it was evaluated, so its apparent ability to identify two agent errors without escalating a correct decision may reflect the sample used to design it; independent validation on new cases would be required before treating the rule as an operational policy.

Finally, the study evaluates reviewability through stored tool calls, feature attributions, and written agent rationales, but does not include a formal user study with fraud analysts. It therefore does not establish whether the generated transcripts improve investigation quality, reduce review time, or support regulatory and organisational audit requirements. The implemented controls should be understood as governance-oriented design elements rather than a complete governance or compliance framework.

## Directions for Further Evaluation

The most important next step is validation on real transaction data containing genuine relational fraud patterns and evaluated under a temporal split. Further work should also compare inductive and transductive graph construction, repeat the model comparisons across multiple splits, and evaluate the agent across larger and naturally imbalanced case samples, multiple models, repeated runs, and alternative retrieval strategies. The escalation policy should be developed on one dataset and tested on another, with separate mechanisms for classifier uncertainty and classifier-agent disagreement. A user study with fraud analysts would also be needed to determine whether the generated evidence and case transcripts improve review quality, reduce investigation time, or support existing case-management and compliance processes.

## Conclusion

This work examined how graph-derived features, an anomaly signal, feature attribution, and a bounded LLM investigation agent contribute within a single fraud analysis pipeline. Three findings stand out. First, after removing a simulator-specific balance shortcut from PaySim, the corrected tabular baseline remained the strongest model on the complete test set, and the additional signals earned their place only in narrower settings: the autoencoder signal ranked fraud best within the fixed subset of intermediate-score cases, and the engineered graph features recovered all injected ring transactions that the baseline detected inconsistently. Second, the investigation agent produced fewer correct verdicts than directly thresholding the classifier it depends on, 39 against 43 on the 60-case sample. In six of its eight disagreements with the classifier it turned a correct decision into an error, and it supplied a coherent written rationale each time. Since every tool available to the agent was derived from the same classifier and feature set, it had no independent evidence with which to outperform that classifier by a wide margin, and the result should temper expectations for investigation agents that only reorganise existing model outputs. Third, a transparent escalation rule based on classifier-agent disagreement identified two of the agent errors in this sample, suggesting that disagreement is a usable signal for routing cases to human review, although the rule requires validation on data it was not designed on.

The practical conclusion is conditional rather than negative. Additional detection signals and investigation agents can contribute, but each needs a defined operational role, an evaluation on the cases it is intended for, and an explicit boundary beyond which its output is not accepted automatically. A reviewable rationale does not certify a correct decision. The most important step toward establishing these results more firmly is evaluation on real transaction data under a temporal split, with the escalation policy developed and tested on separate samples.

## Data and Code Availability

The PaySim dataset is publicly available from the source cited in the references. The implementation, experiment configurations, synthetic-ring generation procedure, and the agent's prompt and tool specifications are available at https://github.com/rahil1303/auditable-fraud-investigation.

## Use of Generative Artificial Intelligence

Generative AI (Claude, Anthropic) was used during the preparation of this manuscript to assist with verification tasks, grammar and prose refinement, and citation style conversion and formatting. The research design, experiments, analysis, and underlying findings are the authors' own. The authors take full responsibility for the content of the final manuscript.